\documentclass[lettersize,journal]{IEEEtran}
\usepackage{amsmath,amsfonts}
\usepackage{algorithmic}
\usepackage{algorithm}
\usepackage{array}
\usepackage{textcomp}
\usepackage{stfloats}
\usepackage{url}
\usepackage{verbatim}
\usepackage{graphicx}
\usepackage{subcaption}
\usepackage{cite}
\usepackage{multirow}
\usepackage{booktabs}
\usepackage{colortbl}
\definecolor{mygray}{gray}{.9}
\definecolor{realred}{RGB}{191,0,64}
\definecolor{realgreen}{RGB}{0,136,124}
\definecolor{realblue}{RGB}{0,124,239}
\usepackage{pifont}
\usepackage{ulem}

\begin{document}

\title{\fontsize{23}{13.5}\selectfont AdaMesh: Personalized Facial Expressions and Head Poses for Adaptive Speech-Driven 3D Facial Animation}


\author{Liyang Chen$^{*}$,~\IEEEmembership{Student Member,~IEEE,}
        Weihong Bao$^{*}$,
        Shun Lei,~\IEEEmembership{Student Member,~IEEE,}
        Boshi Tang, \\
        Zhiyong Wu,~\IEEEmembership{Member,~IEEE,}
        Shiyin Kang,~\IEEEmembership{Member,~IEEE,}
        Haozhi Huang,
        Helen Meng,~\IEEEmembership{Fellow,~IEEE}
        
\thanks{This work is supported by National Natural Science Foundation of China (62076144), Shenzhen Key Laboratory of next generation interactive media innovative technology (ZDSYS20210623092001004) and Shenzhen Science and Technology Program (WDZC20220816140515001, JCYJ20220818101014030). \textit{(Liyang Chen and Weihong Bao contribute equally to this work. Corresponding author: Zhiyong Wu.)}}
\thanks{Liyang Chen, Weihong Bao, Shun Lei, Boshi Tang and Zhiyong Wu are with Tsinghua-CUHK Joint Research Center for Media Sciences, Technologies and Systems, Shenzhen International Graduate School, Tsinghua University, Shenzhen, China (e-mail: \{cly21, bwh21, leis21, tbs22\}@mails.tsinghua.edu.cn; zywu@sz.tsinghua.edu.cn).}
\thanks{Shiyin Kang is with SenseTime Research, Shenzhen, China (e-mail: kangshiyin@sensetime.com).}
\thanks{Haozhi Huang is with XVerse Technology, Shenzhen, China (e-mail: huanghz08@gmail.com).}
\thanks{Helen Meng is with the Department of Systems Engineering and Engineering Management, The Chinese University of Hong Kong, Hong Kong SAR, China (e-mail: hmmeng@se.cuhk.edu.hk).}}

\markboth{IEEE TRANSACTIONS ON MULTIMEDIA, VOL. XX, NO. X, XX 2025}%
{Shell \MakeLowercase{\textit{et al.}}: A Sample Article Using IEEEtran.cls for IEEE Journals}


\maketitle

\begin{abstract}
Speech-driven 3D facial animation aims at generating facial movements that are synchronized with the driving speech, which has been widely explored recently. Existing works mostly neglect the person-specific talking style in generation, including facial expression and head pose styles. Several works intend to capture the personalities by fine-tuning modules. However, limited training data leads to the lack of vividness.
In this work, we propose \textbf{AdaMesh}, a novel adaptive speech-driven facial animation approach, which learns the personalized talking style from a reference video of about 10 seconds and generates vivid facial expressions and head poses. Specifically, 
we propose mixture-of-low-rank adaptation (MoLoRA) to fine-tune the expression adapter, which efficiently captures the facial expression style.
For the personalized pose style, we propose a pose adapter by building a discrete pose prior and retrieving the appropriate style embedding with a semantic-aware pose style matrix without fine-tuning. Extensive experimental results show that our approach outperforms state-of-the-art methods, preserves the talking style in the reference video, and generates vivid facial animation. The supplementary video and code will be available on the project page https://github.com/thuhcsi/AdaMesh.


\end{abstract}
\section{Introduction}
\IEEEPARstart{S}peech-driven 3D facial animation has been widely explored and attracted increasing interest from both academics and industry. This technology has great potential in virtual reality, film production, and game creation. 
Most previous works focus on improving synchronization between speech and lip movements \cite{voca_2019,meshtalk_2021,faceformer_2022,codetalker_2023, sdnerf_2024}, however, they have neglected the personalized talking style, including the facial expressions and head poses. The facial animation with expressive facial expressions and diverse head poses contributes to a more vivid virtual character and grants users with a better interaction experience.

\begin{figure}[t]
    \centering
    \includegraphics[width=0.41\textwidth]{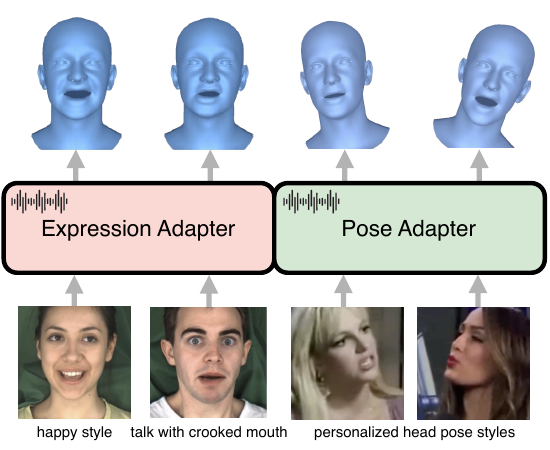}
    \caption{The overview of AdaMesh. The expression adapter and pose adapter generate personalized facial expressions and head poses with the given speech signal and reference talking styles.}
    \label{fig:overview}
\end{figure}
The recent works \cite{mimic_2024,diffposetalk_2024} attempt to capture the speaking style with a style encoder, but the learned facial expression styles are in lack of sufficient richness.
Some works \cite{geneface_2023, imitator_2023, laughtalk_2024} model the person-specific talking style by fine-tuning or adapting modules.
In the real application, only a limited amount of video clips, even shorter than 1 minute, are provided by target users to capture the personalized talking style. These methods thus meet several challenges. 1) The few adaptation data probably induces catastrophic forgetting \cite{catastrophy_2023} for the pre-trained model and easily leads to the overfitting problem \cite{fewshot_zhang_2022}. Specifically for the facial expressions, after the adaptation, the lip synchronization and richness of expressions significantly decrease for the unseen speech inputs. 2) Speech is a weak control signal for the head poses \cite{moglow_2020, facial_2021}. Adaption with few data or learning mappings on such weak signals leads to averaged generation. The predicted head poses are in lack of diversity and have a smaller variation of movements. 3) The facial expressions and head poses are usually modeled by a one-size-fits-all approach \cite{facial_2021, geneface_2023}, neglecting the unique intrinsic characteristics of facial expressions and head poses.

In this work, to address these challenges, we propose \textbf{AdaMesh} as illustrated in Fig. \ref{fig:overview}, an \textbf{adaptive} speech-driven 3D facial animation approach. In consideration of the distinct data characteristics of facial expressions and head poses, we devise two specialized adapters—an expression adapter and a pose adapter—tailored to effectively capture the person-specific talking styles, eschewing the one-size-fits-all approach.

The reference facial expression data provided by users are usually scarce. In an effort to capture expression style, directly fine-tuning the entire model with such limited data often leads to severe overfitting or catastrophic forgetting issues. To tackle these problems, we devise an expression adapter built upon the low-rank adaptation technique, LoRA \cite{lora_2022}, for its high data efficiency in the few-shot model adaptation.
To model the multi-scale temporal dynamics in facial expressions, we extend the vanilla LoRA to the convolution operation and further propose a novel mixture-of-LoRA (MoLoRA) strategy, featured by employing LoRA with different rank sizes.

As for the averaged pose generation problem, we recognize it as a repercussion of generating poses via motion regression \cite{codetalker_2023} and a mismatch of the adapted pose style. Hence we formulate the pose generation task as a generative one, for which we employ a vector quantized-variational autoencoder (VQ-VAE) \cite{vqvae_2018} and a generative pre-trained Transformer (PoseGPT) network \cite{transformer_2017} as the backbone of the pose adapter. 
The head poses are mostly associated with the conveyance of negation, affirmation, and turnaround in semantics. To balance the strong and weak semantic associations in head pose patterns, we propose the retrieval strategy for the pose style adaptation with a novel semantic-aware pose style matrix.

The main contributions can be summarized as follows: 
\begin{itemize}
    \item We propose \textbf{AdaMesh}, which separately models the unique data characteristics of facial expressions and head poses and generates \textbf{personalized} speech-driven 3D facial animation with \textbf{limited adaptation data}.
    \item We are the first to introduce LoRA to the 3D facial animation task and propose MoLoRA to efficiently capture the multi-scale temporal dynamics in facial expressions.
    \item We propose a pose adapter, featured by a semantic-aware pose style matrix and a simple retrieval strategy for diverse pose generation without fine-tuning any parameters.
    \item We conduct extensive quantitative, qualitative and analysis on our AdaMesh. Results show that AdaMesh outperforms other state-of-the-art methods and generates a more vivid virtual character with rich facial expressions and diverse head poses.
\end{itemize}

\section{Related Work}
\subsection{Speech-Driven 3D Facial Animation}
The recent speech-driven 3D facial animation methods can be divided into two categories according to the format of animation. The first category of methods \cite{voca_2019, meshtalk_2021, faceformer_2022, codetalker_2023, imitator_2023} directly predicts vertices of the face from speech and renders the vertices to meshes for visualization. The vertice data is collected through professional 3D scans of human heads \cite{biwi_2010, voca_2019, flame_2017}, which is expensive and time-consuming. The methods trained on such data usually have authentic lip movements. Another category of methods predicts parameters of expressions \cite{emote} or blendshapes \cite{transformers2a,emotalk}. These parameters can be reconstructed from the scanned vertices or 2D face videos through a parametric face model \cite{flame_2017, emotalk}. However, the decreasing quality caused by the reconstruction usually leads to lower speech-lip synchronization.
The latest methods that consider personalized style mostly adopt expression parameters for style modeling, since the scanned vertices are in the neutral style and lack rich expressions. EmoTalk \cite{emotalk} and EMOTE \cite{emote} are related works, which generate emotion-controllable facial animations. EmoTalk adopts unavailable reconstruction pipelines. EMOTE involves a complicated content-emotion disentanglement mechanism. These methods can switch emotions in the training labels but cannot generalize to unseen personalized styles. 
Mimic \cite{mimic_2024} and DiffPoseTalk \cite{diffposetalk_2024} are concurrent works that adopt a style encoder to encode speaking styles, but the generated facial expressions lack sufficient expressiveness.
The most related work is Imitator \cite{imitator_2023}, which generates meshes and leverages the expressions from 2D video for style adaptation.

\subsection{Head Poses Generation}
Existing speech-driven 3D facial animation methods mostly neglect the head poses, since the scanned vertices have no head movements. Some portrait-realistic talking face generation methods attempt to model the head poses. A typical way is to copy the head pose sequence from a real video recording \cite{nvp_2020,geneface_2023,vast_2023}, which can definitely generate realistic head poses. However, these methods disregard the correlation between speech and head pose, and the repetitive and non-engaging head poses create unnaturalness during long periods of interaction. The rapidly growing research topic, gesture generation from speech \cite{taming_2023}, also justifies the importance of predicting motions from speech rather than utilizing a template motion sequence. Another way to predict head poses from speech also attracts many works \cite{zhanghead2007, jia2013headaf, audio2head2021, facial_2021, headtmm_2023, sadtalker_2023, luhead2024}. They have demonstrated the correlation between speech and head pose.
However, these methods can only generate head poses with specific styles, and cannot realize flexible style adaptation. 
Our approach could be regarded as bridging the advantages of the aforementioned two ways. The quantized pose space learned by VQ-VAE ensures realism, while PoseGPT predicts natural but not repetitive poses.

\subsection{Adaptation Strategy}
Efficient adaptation strategies of pre-trained models, including adapter-based \cite{adapter_2017}, prefix-based \cite{prefix_2021}, and LoRA \cite{lora_2022} techniques, have been applied to various downstream applications. These strategies achieve advanced performance without fine-tuning all the model parameters. In the field of image synthesis \cite{sd_2022}, LoRA serves as a plug-and-play extension for the pre-trained model to bring tremendous and specific styles to the images. This work is inspired by this idea and intends to bring the person-specific style to the pre-trained networks that predict facial expressions from speech. Note that we refer to the models proposed in Sec. \ref{sec:expression_adapter} and Sec. \ref{sec:pose_adapter} as the adapters. It is different from the concept of adapter-based strategy \cite{adapter_2017} mentioned above.

\subsection{Remark}
Our work differs from previous works by the enhanced data efficiency in style adaptation and overcomes the challenges of catastrophic forgetting for facial expressions and averaged pose generation. Moreover, compared with the method \cite{laughtalk_2024} that learns specific talking styles on the collected dataset, our work intends to transfer unseen personal talking styles and relies on no large-scale expressive facial animation datasets.
\begin{figure*}[t]
    \centering
    \includegraphics[width=0.98\textwidth]{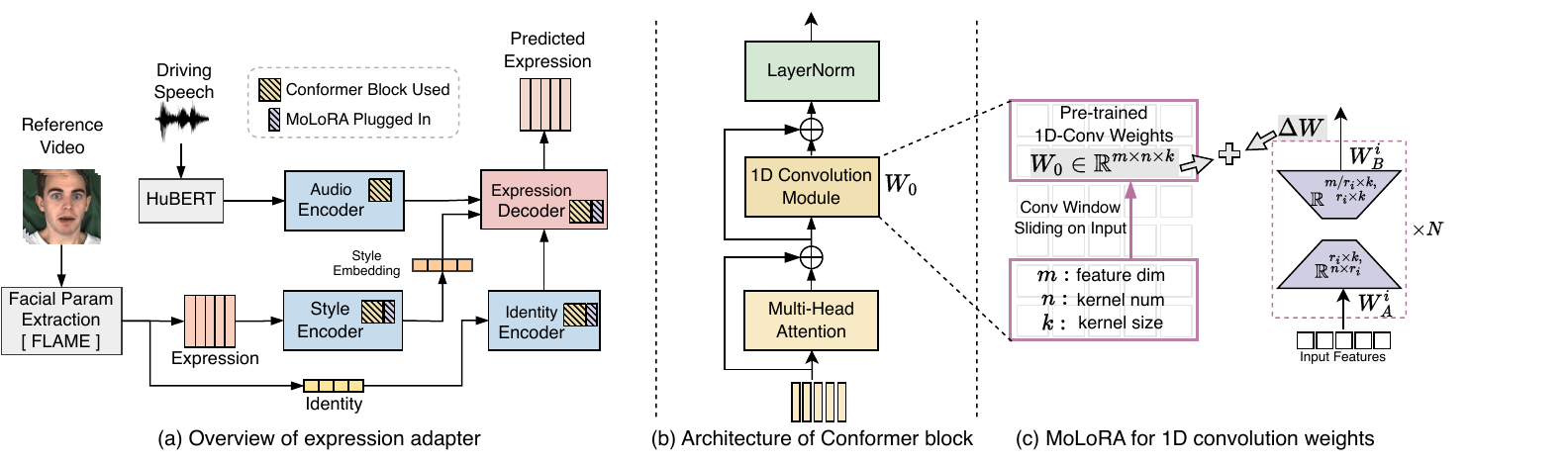} 
    \caption{The overview of expression adapter. (a) The patches on each encoder and decoder denote the Conformer blocks are used and MoLoRA parameters are added to the pre-trained modules after adaptation. (b) A brief illustration of Conformer \cite{conformer_2021} (c) Illustration of 1D-Convolution weights and MoLoRA weights applied on the input features. MoLoRA combines $N$ LoRAs with different rank sizes of $r_i$. MoLoRA parameters are added to the convolution and linear layers in the Conformer blocks of the encoders and decoder to efficiently learn the expression style from the reference data.}
    \label{fig:figure1}
\end{figure*}
\section{AdaMesh}
\subsection{Overview}
In this section, we propose AdaMesh, a data-efficient speech-driven 3D facial animation approach tailored for generating personalized facial expressions and head poses.
Facial expressions and head poses have distinct and unique data characteristics.
The facial expression are strongly correlated with speech, and the basic facial movements can be accurately predicted from the speech content. The expression style can be viewed as a deviation signal superimposed on these movements, resulting in the display of different emotions. The head poses exhibit strong personal styles and convey specific semantics \cite{zhanghead2007, audio2head2021}. It can be regarded as a periodic signal superimposed with speech-related peaks and troughs. Therefore, we employ two independent adapters to separately capture expression and head pose styles.
Given the facial expression sequence and head pose sequence extracted from a reference video, together with a driving speech, the expression adapter (Sec. \ref{sec:expression_adapter}) is adapted with the specific expression style and outputs rich facial expressions; while the pose adapter (Sec. \ref{sec:pose_adapter}) derives a semantic-pose aware style matrix, and outputs diverse head poses with VQ-VAE and PoseGPT. The generated facial expressions and head poses are combined by a parametric face model \cite{flame_2017} and rendered to mesh images.

\textbf{Facial Parameter Representation.} We adopt the morphable head model FLAME \cite{flame_2017, spectre_2023} as 3D face representation. FLAME is described by a combination function $M(\beta, \delta, \psi)$, that takes coefficients describing identity $\beta \in \mathbb{R}^{100}$, expression $\delta \in \mathbb{R}^{50}$, and pose $\psi \in \mathbb{R}^{3k+3}$ ($k=4$ joints) and returns 5023 vertices of xyz dimensions. The 3D vertices can be further rasterized as 2D mesh. More specifically for this work, $\delta$ and the last 3 dimensions of $\psi$ are for facial motion, and the other dimensions of $\psi$ for head pose.

\subsection{Expression Adapter}
\label{sec:expression_adapter}
Directly fine-tuning the entire model for expression prediction may lead to forgetting previously learned knowledge, \normalem\emph{i.e.}, how to generate speech-synced lip movements in this task. To achieve efficient adaptation for facial expressions, we pre-train the expression adapter to learn general and person-agnostic information that ensures lip synchronization and then optimize the MoLoRA parameters to equip the expression adapter with a specific expression style.
As shown in Fig. \ref{fig:figure1} The expression adapter is composed of an audio encoder, an identity encoder, a style encoder, an expression decoder, and MoLoRA to fine-tune convolutional and linear layers in the decoder and the encoders except the audio encoder.
\begin{figure}[t]
    \centering
    \includegraphics[width=0.45\textwidth]{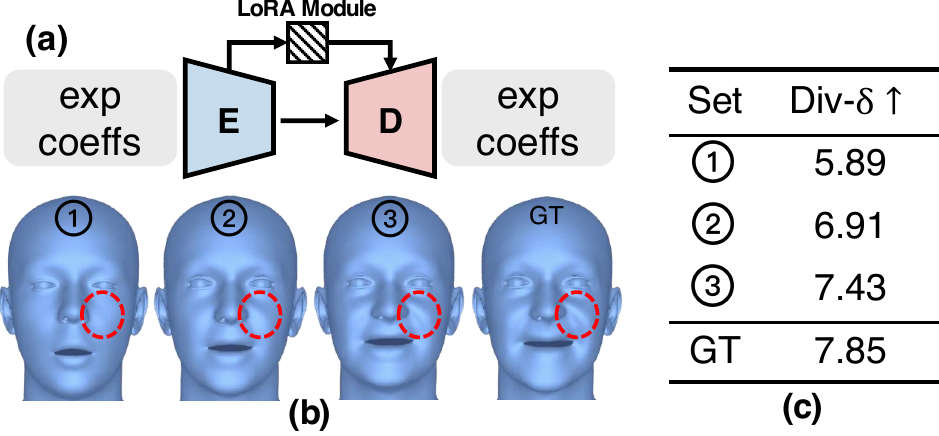}
    \caption{(a) Architecture of the auto-encoder for expression reconstruction. (b) Reconstructed meshes using different rank sizes. (c) Diversity-expression scores for different rank sizes. \ding{172}\ding{173}\ding{174} denote rank size 16, 64 and 128.}
    \label{fig:lora_vis}
\end{figure}

\textbf{Audio Encoder \& Identity Encoder.} 
To obtain robust semantic information from speech, the self-supervised speech representation model HuBERT \cite{hubert_2021} is leveraged to extract features from raw speech. These speech features are then sent into three Conformer layers \cite{conformer_2021}.
Previous methods usually represent the identity with the one-hot label \cite{codetalker_2023,imitator_2023}, leading to the bad generalization on unseen speaker identity. We utilize the identity parameter extracted by the parametric face model \cite{flame_2017, deca_2021}. The identity sequence in a data sample is averaged along the time dimension to represent the speaker identity and then sent into three convolutional layers with conditional layer normalization \cite{adaspeech_2021}.

\textbf{Style Encoder and Expression Decoder.}
To provide the talking style that cannot be directly predicted from the semantic speech information, the expression sequence is utilized as a residual condition to the expression prediction. The sequence is compressed to a compact style embedding after two Conformer layers and then concatenated with the encoded speech and identity features into the decoder. 
The decoder also consists of three Conformer layers, which are stacked in a residual manner \cite{soundstream_2021} rather than cascadation to realize coarse-to-fine prediction.

\begin{figure*}[t]
    \centering
    \includegraphics[width=0.9\textwidth]{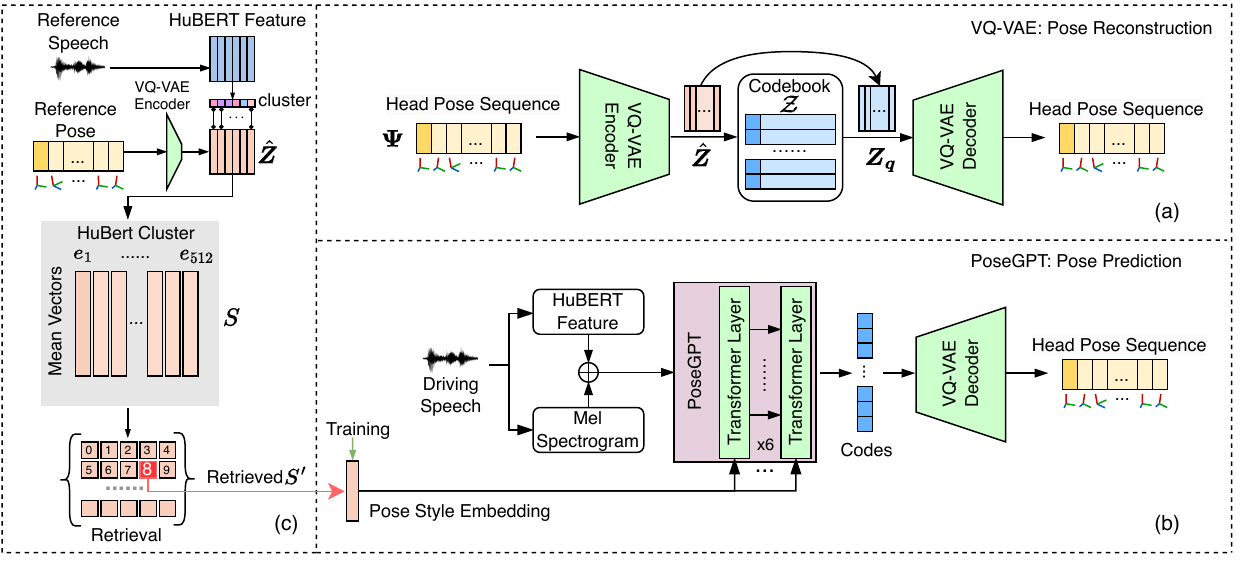}
    \caption{The overview of pose adapter. (a) Training of VQ-VAE. (b) PoseGPT. (c) The derivation of the semantic-aware pose style matrix and the retrieval strategy. In the training of the PoseGPT, the pose style embedding is the assigned one-hot label for each sample. $S$ denotes the semantic-aware pose style matrix. More details about the training and inference can be found in the supplementary materials.}
    \label{fig:figure2}
\end{figure*}
\textbf{MoLoRA.}
We focus on the low-rank training techniques and propose MoLoRA to equip the expression adapter with the inductive bias of new data. As it is known, the vanilla LoRA \cite{lora_2022} injects trainable rank decomposition matrices into each linear layer of the Transformer architecture. Since the convolutional layer can be viewed as multiple linear transformations with sliding receptive windows, we extend LoRA to the convolution layer for more flexible trainable parameters. 
The facial expressions contain multi-scale temporal dynamics, where the coarse-grained global features (\textit{e.g.,} emotion) and fine-grained facial details (\textit{e.g.,} speech-sync lip movements, muscle wrinkles) inherent in the expression sequence.
As shown in Fig. \ref{fig:lora_vis}, we reconstruct the expression coefficients with a simple auto-encoder \cite{autoencoder_2022} and inject the vanilla LoRA modules with different rank sizes for expression observation. We empirically found that the LoRA with a small rank size tends to preserve coarse features and a bigger rank size captures more fine-grained dynamics, especially around the cheek region. The quantitative metric in Fig. \ref{fig:lora_vis}(c) also proves that a bigger rank size results in higher diversity. The latest works \cite{talklora_2024,metaloraanimation_2024} demonstrate that the LoRA strategy can efficiently capture facial motion details while some studies \cite{relora_2023, lora_2022} show that combining LoRA of different ranks can achieve more efficient training and better performance in large language models. The results in Fig. \ref{fig:lora_vis} are in line with the observations of these works.

Hence we propose MoLoRA by mixing $N$ LoRAs with different rank sizes $r_{i}$ for expression style modeling. 
Supposing the pretrained weight matrix of 1D convolution $W_0 \in \mathbb{R}^{m \times n \times k}$, we define the low-rank product $\Delta W^i = W_{B}^{i}W_{A}^{i}$, where $W_{B}^{i} \in \mathbb{R}^{m/r_i \times k, r_i \times k}$ and $W_{A}^{i} \in \mathbb{R}^{r_i \times k, n \times r_i}$. The weights $W$ after adaptation merge the pre-trained and MoLoRA weights, which can be defined as: 
\begin{align}
    \label{eq: MoLoRA}
    W = W_0 + \Delta W = W_0 + \sum_{i=0}^{N}W_{B}^{i}W_{A}^{i}.
\end{align}
We perform the same reshape operation as the vanilla LoRA \cite{lora_2022}, converting the dimension $\Delta W^i$ as $m/r_i \times k \times n \times r_i \rightarrow{m \times n \times k}$, which is consistent with the pre-trained weights $W_0$. There is a concurrent work \cite{relora_2023} that investigates the efficiency of LoRA. However, it stacks multiple LoRA of a single rank size, while MoLoRA focuses on modeling the multi-scale characteristic in facial expressions.

\textbf{Procedure of Expression Adapter.}
The pipeline of the expression adapter is listed in Alg. \ref{alg:expression_adapter}. Note that the inference stage has a slight difference with the training. The style embedding is obtained from another expression sequence rather than the ground truth. However, this mismatch has no impact on the generation, since the style embedding learns a global talking style for the expression sequence rather than the semantic speech information.
\begin{algorithm}
    \caption{Pipeline of Expression Adapter.}
    \begin{algorithmic}
        \STATE \textbf{Pretrain:} Train the Expression Adapter $\theta_{E}$ with source training data. The expression sequence for the style encoder is the same with the prediction of the expression adapter.
        \STATE \textbf{Adaptation:} Update all low-rank matrices $\theta_{MoLoRA}$ for convolution and linear layers in the expression adapter with the adaptation data.
        \STATE \textbf{Inference:} Merge $\theta_{E}$ and $\theta_{MoLoRA}$. Given a driving speech, and an expression sequence from the adaptation data as the input for the style encoder, the expression adapter generates expressions that synchronize with the speech and reenact the facial style of the adaptation data.
    \end{algorithmic}
    \label{alg:expression_adapter}
\end{algorithm}

\textbf{Loss Function.} We adopt the simple mean square error (MSE) loss at both the pretrain and adaptation stages for the expression adapter. The model is optimized for 300k steps and 30 steps at each stage. 

\subsection{Pose Adapter}
\label{sec:pose_adapter}
To overcome the averaged generation problem in pose style adaptation, we propose the pose adapter and formulate the adaptation as a simple but efficient retrieval task instead of fine-tuning modules. As illustrated in Fig. \ref{fig:figure2}, we first learn a discrete latent codebook $\mathcal{Z}=\{\boldsymbol{z_k} \in \mathbb{R}^{d_{\boldsymbol{z}}}\}_{k=1}^{M}$  for head poses with the novel sequence-encoding VQ-VAE \cite{bailando_2022}, and then predicts the discrete head pose codes from the driving speech with PoseGPT. The VQ-VAE training requires no speech input, so various video datasets of a larger amount can be utilized. This design enables the coverage of a wide range of pose styles, which is well-suited for learning a realistic pose manifold \cite{learn2listen_2022}.
A semantic-aware pose style matrix is derived for each pose sample in the training sample to create a database. We retrieve the nearest matrix in this database as the adapted style matrix in the adaptation stage. The simple retrieval strategy for adaptation is based on the assumption that the possible pose styles have been covered in the VQ-VAE learning with the large-scale dataset. 

\textbf{Discrete Pose Space.} Given head pose sequence in continuous domain $\boldsymbol{\Psi}=[\boldsymbol{\psi}_1, \ldots, \boldsymbol{\psi}_T] \in \mathbb{R}^{T \times 3}$, the VQ-VAE encoder, which consists of 1D convolutional layers, firstly encodes it into context-aware features $\hat{\boldsymbol{Z}} \in \mathbb{R}^{T' \times d_{\boldsymbol{z}}}$, where $T'=T/w$ denotes the frame numbers of the downsampled features.  Then, we obtain the quantized sequence
$\boldsymbol{Z_q}$ via an element-wise quantization function $Q$ that maps each element in $\hat{\boldsymbol{Z}}$ to its closest codebook entry:
\begin{equation}
    \boldsymbol{Z_q}=Q(\hat{\boldsymbol{Z}}):=(\underset{\boldsymbol{z_k} \in \mathcal{Z}}{\arg \min }\|\hat{\boldsymbol{z}}_t-\boldsymbol{z}_k\|_{2}) \in \mathbb{R}^{T' \times d_{\boldsymbol{z}}}.
\end{equation}
$\boldsymbol{Z_q}$ is then reconstructed to continuous poses by VQ-VAE decoder, which consists of three 1D convolutional layers.

\textbf{PoseGPT.} With the learned discrete pose space, we build a PoseGPT network \cite{transformer_2017, bailando_2022} that maps the driving speech and pose style embedding into the discrete codes of the target poses. 
The HuBERT feature and mel spectrogram of the driving speech are utilized for speech representation to consider both speech semantics and speech prosody (\textit{e.g.,} stress, tone) \cite{transformers2a} information. 
We assign each pose sample in the training dataset with a one-hot label as the pose style embedding. The speech representation and pose style embedding are sent into PoseGPT, which consists of six cross-conditional Transformer layers \cite{transformer_2017}. 
PoseGPT predicts discrete codes autoregressive by selecting the output with the highest probability at each step. The predicted codes are finally sent into the VQ-VAE decoder to obtain the continuous head pose sequence. We train  PoseGPT using the teacher-forcing training scheme with the cross-entropy loss.

\textbf{Semantic-Aware Pose Style Matrix.} Given a speech utterance of $T$ frames and the corresponding head pose sequence, HuBERT \cite{hubert_2021} assigns each speech frame into one of 512 clusters, while the VQ-VAE encoder transforms the pose sequence into the latent sequence $\hat{\boldsymbol{Z}}$. Thus, each frame in $\hat{\boldsymbol{Z}}$ is labeled with a cluster which contains the speech semantic information. We calculate the semantic-aware pose style matrix $S \in \mathbb{R}^{512 \times d_z}$ for each pose sequence as:
\begin{equation}
    S_j = \sum_{i=1}^{T'} \hat{\boldsymbol{Z}}_i \cdot \kappa(L_i, j) / \sum_{i=1}^{T'} \kappa(L_i, j), j=[1,\dots, 512],
\end{equation}
where $L_i$ denotes the cluster label for $i$ frame of $\hat{\boldsymbol{Z}}$ and $\kappa$ is the Kronecker symbol that outputs 1 when the inputs are equal else 0. 
This equation aggregates all the poses which have the same semantic unit of HuBERT.
Since the clustering step in HuBERT \cite{hubert_2021} captures the semantic information in acoustic units and the VQ-VAE encoder models the temporal pose style information, the derived  $S$ represents the pose style that considers semantics.

\textbf{Procedure of Pose Adapter.}
The pipeline of the pose adapter is listed in Alg. \ref{alg:pose_adapter}. Note that the adaptation stage involves no parameter updating.
\begin{algorithm}
    \caption{Pipeline of Pose Adapter.}
    \begin{algorithmic}
        \STATE \textbf{Pretrain:} For each pose sequence in the source training dataset, assign it with a one-hot pose style embedding, and calculate its semantic-aware pose style matrix with its corresponding speech. Establish the database for all pose style matrices and their paired pose style embedding. Train the VQ-VAE and PoseGPT with the source training dataset.
        \STATE \textbf{Adaptation:} Given an unseen pair of reference speech and pose sequence, calculate the semantic-aware pose style matrix for the reference data. Find the nearest pose style matrix in the established database by computing the L1 distance, and adopt the corresponding pose style embedding for inference.
        \STATE \textbf{Inference:} Given a driving speech, and utilizing the retrieved pose style embedding, the pose adapter generates continuous head poses with the pre-trained PoseGPT and the VQ-VAE decoder.
    \end{algorithmic}
    \label{alg:pose_adapter}
\end{algorithm}

\textbf{Loss Function.} The loss function of VQ-VAE is:
\begin{equation}
    \mathcal{L}_{VQ} = \mathcal{L}_{rec}(\hat{\boldsymbol{\Psi}}, \boldsymbol{\Psi}) + \left\| \text{sg}[\hat{\boldsymbol{z}}] - \boldsymbol{z_q} \right\| + \gamma \left\| \boldsymbol{z} - \text{sg}[\boldsymbol{z_q}] \right\|
\end{equation}
where $\hat{\boldsymbol{\Psi}}$ and $\boldsymbol{\Psi}$ denote predicted and ground truth pose sequences, and $\text{sg}[\cdot]$ denotes the operation of stop gradient. $\mathcal{L}_{rec}(\hat{\boldsymbol{\Psi}}, \boldsymbol{\Psi})$ can be formulated as:
\begin{align}
    \mathcal{L}_{\text {rec }}(\hat{\boldsymbol{\Psi}}, \boldsymbol{\Psi}) &= \left\| \hat{\boldsymbol{\Psi}}-\boldsymbol{\Psi} \right\|_{1} + \alpha_{1} \left\| \hat{\boldsymbol{\Psi}}^{\prime}-\boldsymbol{\Psi}^{\prime} \right\|_{1} \\ \notag
    &+ \alpha_{2} \left\| \hat{\boldsymbol{\Psi}}^{\prime \prime}-\boldsymbol{\Psi}^{\prime \prime} \right\|_{1}
\end{align}
where $\boldsymbol{\Psi}^{\prime}$ and $\boldsymbol{\Psi}^{\prime \prime}$ represent the $1^{st}$-order (velocity) and $2^{nd}$-order (acceleration) partial derivatives of pose sequence $\boldsymbol{\Psi}$ on time. The velocity and acceleration loss items ensure the generation of smooth head poses and prevent abnormal jitters, which is commonly used in previous works \cite{nvp_2020,bailando_2022}.
$\alpha_{1}$, $\alpha_{2}$ and $\gamma$ denote trade-off coefficients.
The PoseGPT is optimized with the MSE loss. To expedite convergence and align with the inference process, we employ a teacher-forcing strategy to train PoseGPT during the early training and gradually transit to the autoregressive training scheme.
\section{Experiments}
\begin{figure*}[!t]
    \centering
    \includegraphics[width=1.0\textwidth]{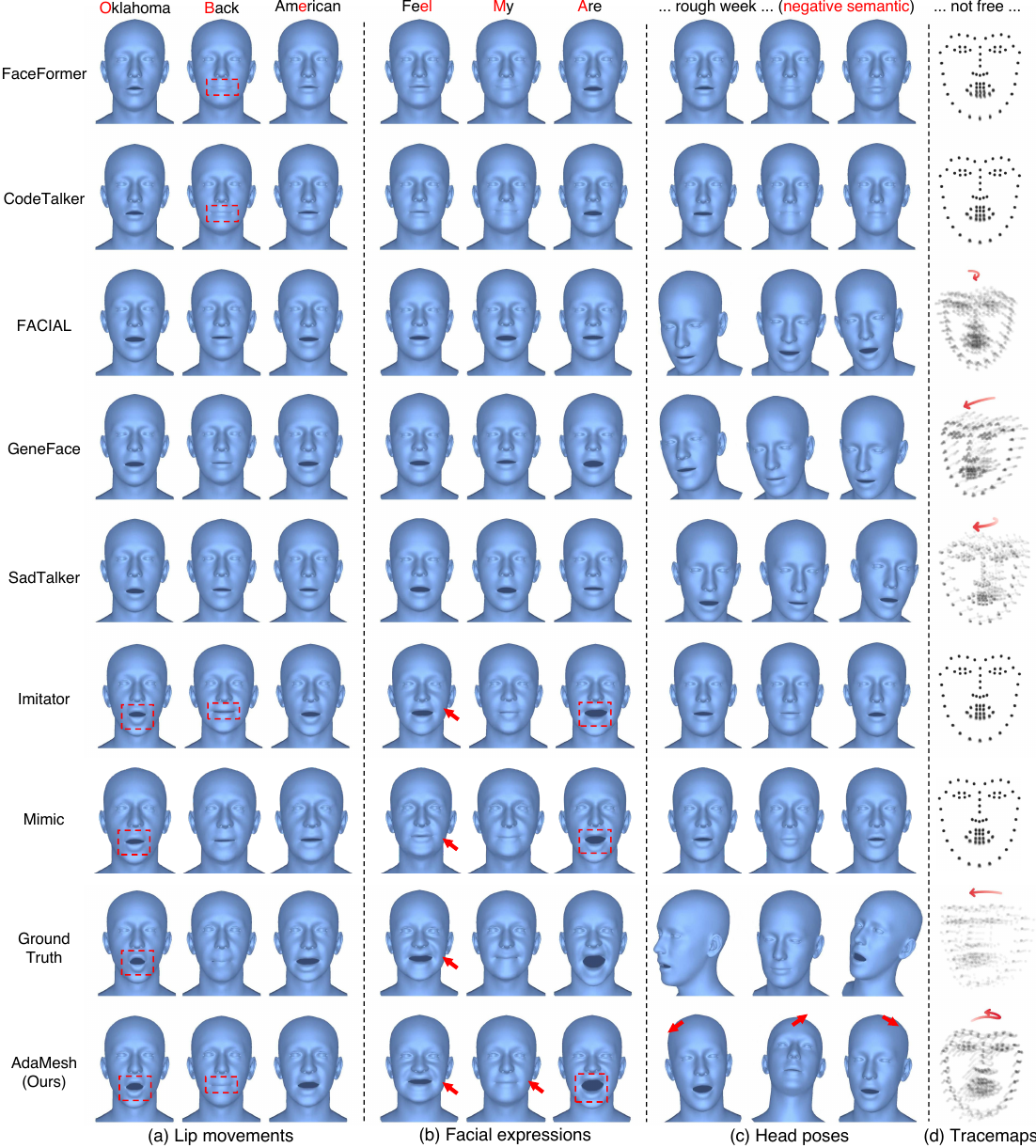} 
    \caption{Qualitative comparison with different methods. (a) shows lip movements on Obama dataset with neutral talking style. (b) is for observation of personalized facial expressions on the emotional MEAD dataset. (c) and (d) show head poses and corresponding landmark tracemaps on the VoxCeleb2-Test dataset. The first row displays the words or sentences that these frames are pronouncing.}
    \label{fig: figure3}
\end{figure*}
\textbf{Dataset.} Due to the scarcity of 3D animation datasets with personalized talking styles, we utilize the 2D audio-visual talking face datasets. For expression adapter pre-train, we utilize Obama weekly footage \cite{synobama_2017} with a neutral talking style. For pose adapter pre-train, VoxCeleb2 \cite{vox2_2018} is utilized, which contains about 1M talking-head videos of 6112 speakers with diverse and natural head poses. In the adaptation stage, we introduce MEAD \cite{mead_2020} for the expression adapter, since MEAD consists of videos featuring 60 actors with 7 different emotions. We utilize VoxCeleb2-Test dataset for the adaptation of the pose adapter, and the speakers are unseen in pre-train.

\textbf{Dataset Processing.} We resample 2D videos as 25fps. 
To derive 3D data from 2D data, we employ the state-of-the-art face reconstruction method, SPECTRE \cite{spectre_2023}, to obtain FLAME coefficients. Scanned data offers more precise details in lip and forehead movement. The average error of extracted 3D data \cite{deca_2021, spectre_2023}, compared to scanned data, is around 1-2mm. Considering the average width and length of a face are approximately 130mm and 200mm, such an error is not easily perceptible. 
To further enhance the fidelity of the reconstructed 3D data, we optimize SPECTRE network for each video clip, rather than using a pre-trained face model for inference directly.
Given the extracted parameters of identity, expression, and head pose, we can compute the 3D vertices of the per-frame meshes. For the audio track, we downsample the raw wave as 16kHz and adopt the base HuBERT model \cite{hubert_2021} to extract speech features. Since the frame rates of video and audio features are different, we interpolate the audio features and align them to the same length \cite{voca_2019}.
\begin{table*}[t]    
    \centering
    \small
    \renewcommand{\arraystretch}{1.0}  
    \caption{Quantitative evaluation and user study results. User study scores are with 95\% confidence intervals. Exp-Richness value reveals the percentages of users who prefer the compared method other than AdaMesh in A/B tests. Value lower than 50\% indicates that users prefer AdaMesh (Ours). Collab: two adapters collaborate.}
    \label{tab:table1}
    \resizebox{1\textwidth}{!}{
        \begin{tabular}{lcccccccccc}
            \toprule
            \multirow{2}{*}{} & \multicolumn{3}{c}{Facial Expression} & \multicolumn{4}{c}{Head Pose} & \multicolumn{3}{c}{User Study}\\
            \cmidrule(lr){2-4} \cmidrule(lr){5-8} \cmidrule(lr){9-11} 
            & LVE$\downarrow$ & EVE$\downarrow$ & Div-{$\boldsymbol{\delta}$}$\uparrow$ & FID$\downarrow$ & LSD$\uparrow$ & Div-{$\boldsymbol{\psi}$}$\uparrow$ & FSD$\downarrow$ & Lip-Sync &Pose-Natural &Exp-Richness\\
            \midrule
            FaceFormer &4.67 &3.46 &6.42 &1.05 &-- &-- &-- &4.05$\pm$0.15 &1.46$\pm$0.10 &0.80\%\\
            Codetalker &4.43 &3.30 &6.48 &1.05 &-- &-- &-- &4.10$\pm$0.14 &1.42$\pm$0.11 &0.80\%\\
            FACIAL &3.44 &2.48 &6.32 &1.68 &0.69 &1.50 &30.65 &3.88$\pm$0.21 &3.85$\pm$0.20 &11.7\%\\
            GeneFace &3.54 &2.47 &6.56 &1.38 &0.70 &1.49 &34.13 &4.01$\pm$0.18 &3.79$\pm$0.17 &12.3\%\\
            SadTalker &4.80 &3.62 &6.42 &1.38 &0.53 &1.11 &62.40 &3.88$\pm$0.14 &3.82$\pm$0.15 &11.5\% \\
            Imitator &3.15 &2.33 &5.97 &1.05 &-- &-- &-- &3.98$\pm$0.19 &1.42$\pm$0.11 &29.8\%\\
            {Mimic} &{3.03} &{2.30} &{6.67} &{1.05} &{--} &{--} &{--} &{4.05$\pm$0.18} &{1.42$\pm$0.11} &{40.2\%}\\
            \hline
            Ground Truth &0 &0 &7.85 &-- &1.37 &3.01 &-- &4.30$\pm$0.14 &4.12$\pm$0.23 &41.3\% \\
            AdaMesh (Ours) &\textbf{2.91} &\textbf{2.25} &\textbf{7.10} &\textbf{0.90} &\textbf{0.87} &\textbf{1.57} &\textbf{13.89} &\textbf{4.12$\pm$0.19} &\textbf{4.22$\pm$0.11} &--\\
            {AdaMesh (Collab)} &{2.93} &{2.25} &{7.08} &{1.66} &{0.55} &{1.21} &{57.43} &{4.10$\pm$0.18} &{3.55$\pm$0.20} &{48.3\%}\\
            \bottomrule
        \end{tabular}
    }
\end{table*}

\textbf{Baseline Methods.} We compare AdaMesh with several state-of-the-art methods. 1) FaceFormer \cite{faceformer_2022}, which first introduces autoregressive Transformer-based architecture for speech-driven 3D facial animation. 2) CodeTalker \cite{codetalker_2023}, which is the first attempt to model facial motion space with discrete primitives. 
3) FACIAL \cite{facial_2021}, which focuses on predicting dynamic face attributes, including head poses. 
4) GeneFace \cite{geneface_2023}, which proposes a domain adaptation pipeline to bridge the domain gap between the large corpus and the target person video. It can also generate realistic head poses with a template pose sequence.
5) SadTalker \cite{sadtalker_2023}, which learns the realistic 3D motion coefficient and generalize to various images.
6) Imitator \cite{imitator_2023}, which learns identity-specific details from a short input video and produces facial expressions matching the personalized speaking style.
7) Mimic \cite{mimic_2024}, which learns disentangled representations of the speaking style and content from facial motions by building two latent spaces. FACIAL, GeneFace and SadTalker are photo-realistic talking face methods. We therefore reconstruct 3D meshes \cite{deca_2021} from the generated videos for FACIAL and GeneFace, and convert the BFM \cite{deep3d_2019} to FLAME \cite{flame_2017} format\footnote{https://github.com/TimoBolkart/TF\_FLAME} for SadTalker.
\section{Results}
\subsection{Qualitative Results}
To reflect the perceptual quality, we conduct qualitative evaluation from three aspects as shown in Fig. \ref{fig: figure3}. Best viewed with the demo video on the AdaMesh project page.

\textbf{Speech-Lip Synchronization.} We illustrate three keyframes speaking the vowels and consonants. The visual results show that our approach generates more accurate lip movements than other methods. AdaMesh has more pronounced mouth open when speaking /ou/ and /ei/. FaceFormer and CodeTalker close their mouth tighter on /b/ than other methods, since they are directly trained on the 3D scanned vertices. Other methods are trained or fine-tuned on the reconstructed data from 2D video, which are less precise than the scanned data.

\textbf{Facial Expression Richness.} From Fig. \ref{fig: figure3}(b), it can be observed that our approach preserves the expressive facial style and obviously gains richer facial details, especially around the cheek and lip regions. AdaMesh even gains more expressive details and accurate lip movements than the ground truth in the supplementary video. The MoLoRA parameters in AdaMesh act as a plugin of facial style for the pre-trained model, and the pre-trained model on the Obama dataset has guaranteed lip synchronization. {It can be observed that the introduction of MoLoRA significantly enhances the richness of facial expressions without compromising the accuracy of lip movements. Since the audio-lip sync performance of the adaptation data is typically close to the average of the pre-trained dataset, the MoLoRA strategy focuses on capturing the differences, particularly the expression style, between the adaptation dataset and the pre-trained dataset.}


\textbf{Head Pose Naturalness.} As shown in Fig. \ref{fig: figure3}(c), we illustrate frames in a sentence that contains evident emotion. A sentence that conveys negative semantics is accompanied by more evident head poses, which is consistent with the FSD result in the quantitative evaluation. Our approach produces head movements with a greater range of amplitudes, and the results are closer to the ground truth. Following previous works \cite{facial_2021, sadtalker_2023}, we also plot the landmark tracemaps for a sentence that conveys negative semantics. The head poses with more degrees of freedom confirm the diversity generation of AdaMesh.

\textbf{User Study.} We conduct a user study to measure the perceptual quality of these methods. Fifteen users with professional English ability participate in this study. For lip synchronization and pose naturalness, we adopt mean-opinion-score (MOS) tests to ask users to score each video from one (worst) to five (best). Our approach has achieved the highest scores on these two aspects. For expression richness, we adopt A/B tests to ask users to select a better video between AdaMesh and another method. Our approach has significantly outperformed other methods, including the ground truth. This is attributed to the image jittering brought by \cite{deca_2021, flame_2017}. AdaMesh has filtered out these disturbances in the pre-training of the expression adapter. 
\begin{table}[tbp]    
    \centering
    \caption{Comparsion results on VOCASET. FT denotes finetuning with few VOCASET data.}
    \label{tab:vocaset}
    \begin{tabular}{lcc}
        \toprule
        Method   & LVE (mm) $\downarrow$  & Trained on VOCASET \\ 
        \midrule
        FaceFormer                 & 4.3011 & Yes       \\
        CodeTalker    & 4.2902 & Yes      \\
        Ours w/ FT & \textbf{4.2237} & 30s adaptation      \\
        Ours wo/ FT           & 4.2933  & No      \\
        \bottomrule
    \end{tabular}
\end{table}
\subsection{Quantitative Evaluation}
\textbf{Evaluation Metric.} To measure lip synchronization, we calculate the lip vertex error (LVE), which is widely used in previous methods \cite{faceformer_2022, imitator_2023}. We follow \cite{emotalk} to calculate the emotional vertex error (EVE) to evaluate if the method can capture the personalized emotional talking style on MEAD dataset \cite{mead_2020}. The diversity scores \cite{bailando_2022,vast_2023} for expressions (Div-{$\boldsymbol{\delta}$}) and poses (Div-{$\boldsymbol{\psi}$}) are separately computed to evaluate if the methods generate vivid and diverse motions. The landmark standard deviation (LSD) \cite{facial_2021, codetalker_2023} measures the pose dynamics. The FID score \cite{bailando_2022,taming_2023} estimates the head pose realism. To evaluate if the head poses are semantically aligned with the speech, we calculate the FSD score by computing the FID scores which correspond to the same HuBERT units and then averaging these scores \cite{hubert_2021}.

\textbf{Evaluation Results.} The results are shown in Tab.\ref{tab:table1}. For facial expressions, it can be observed that our approach obtains lower LVE than other methods, which indicates AdaMesh has more accurate lip movements. FaceFormer and CodeTalker have poorer performance, since the LVE score is calculated on MEAD and Obama datasets, and these two methods have poor generalization ability to unseen speaker identity. AdaMesh also achieves the lowest EVE. It suggests that our approach can efficiently capture the facial style in the adaptation. A higher diversity score for expressions confirms that our approach generates more vivid facial dynamics. For head poses, FaceFormer, CodeTalker, and Imitator cannot generate head poses. GeneFace incorporates a real head pose sequence for presentation and ~FACIAL learns over-smoothed poses. The FID scores show that our approach generates more realistic head poses, while LSD and diversity scores confirm our approach generates head poses with higher diversity. AdaMesh also gains the lowest FSD score, suggesting that the generated head poses are semantic aware.
{To further explore whether the collaboration of two adapters can collectively contribute to the task, we combine the expression and pose adapters, referring it as AdaMesh (Collab). It adopts the MoLoRA strategy to capture speaking styles, and simultaneously generates expression and pose parameters. Compared with AdaMesh (Ours), AdaMesh (Collab) shows a slight decrease in performance related to expression and a significant decrease in pose-related metrics. This is ascribed to the collaborative generation disregarding the intrinsic data characteristics of head poses. It substantiates the rationale for employing separate adapters to capture speaking styles.}
\begin{table}[tbp]    
    \centering
    \small
    \caption{Ablation study for expression adapter.}
    \label{tab:table2}
    \begin{tabular}{lccc}
        \toprule
        Setting   & LVE$\downarrow$  & EVE$\downarrow$  & Div-{$\boldsymbol{\delta}$}$\uparrow$ \\ 
        \midrule
        Ours                 & \textbf{2.91} & \textbf{2.25} & \textbf{7.10}       \\
        w/o style encoder    & 3.12 & 2.34 & 6.33      \\
        w/o identity encoder & 3.05 & 2.29 & 6.98      \\
        w/o MoLoRA           & 3.10  & 2.39 & 6.21      \\
        \bottomrule
    \end{tabular}
\end{table}
\begin{table}[tbp]    
    \centering
    \small
    \caption{Ablation study for pose adapter.}
    \label{tab: abla_pose}
    \begin{tabular}{lcccc}
        \toprule
        Setting   & FID$\downarrow$  & LSD$\uparrow$  & Div-{$\boldsymbol{\psi}$}$\uparrow$ &FSD$\downarrow$ \\ 
        \midrule
        Ours                 & \textbf{0.90} & \textbf{0.87} & \textbf{1.57} &\textbf{13.89}      \\
        w/o style matrix    & 1.01 & 0.69 & 0.69 &19.88      \\
        w/o HuBERT feature & 0.91 & 0.07 & 0.15 & 26.73      \\
        w/o VQ-VAE           & 0.91  & 0.05 & 0.02 & 22      \\
        \bottomrule
    \end{tabular}
\end{table}

\textbf{Comparison on VOCASET.} Our approach is designed for personalized facial expressions and head poses. However, VOCASET \cite{voca_2019} has no explicit talking styles or head poses. To evaluate the generalization ability of our approach, we convert the VOCASET vertices to expression coefficients for inference. 
We test our approach on two settings. 1) Ours w/o FT: we directly run the inference pipeline of the expression adapter with the unseen speech in VOCASET without fine-tuning any parameters. The converted expressions are sent into the style encoder of the expression adapter. 
2) Ours w/ FT: we fintune the MoLoRA parameters of the expression adapter with the converted expressions and speech data. Since the converted expressions are not as accurate as the original vertices, we train FaceFormer \cite{faceformer_2022} and CodeTalker \cite{codetalker_2023} on the converted data for a fair comparison.
The lip vertex error (LVE) calculated in Tab.\ref{tab:vocaset} indicates that our approach achieves competitive performance, although AdaMesh is not directly trained on VOCASET. This confirms the generalization ability of our approach. When the MoLoRA parameters in the expression adapter are updated with only 30 seconds of data, our approach gains better performance than other methods.
\begin{figure}[t]
  \centering
  \begin{subfigure}{0.48\linewidth}
    \centering
    \includegraphics[scale=0.29]{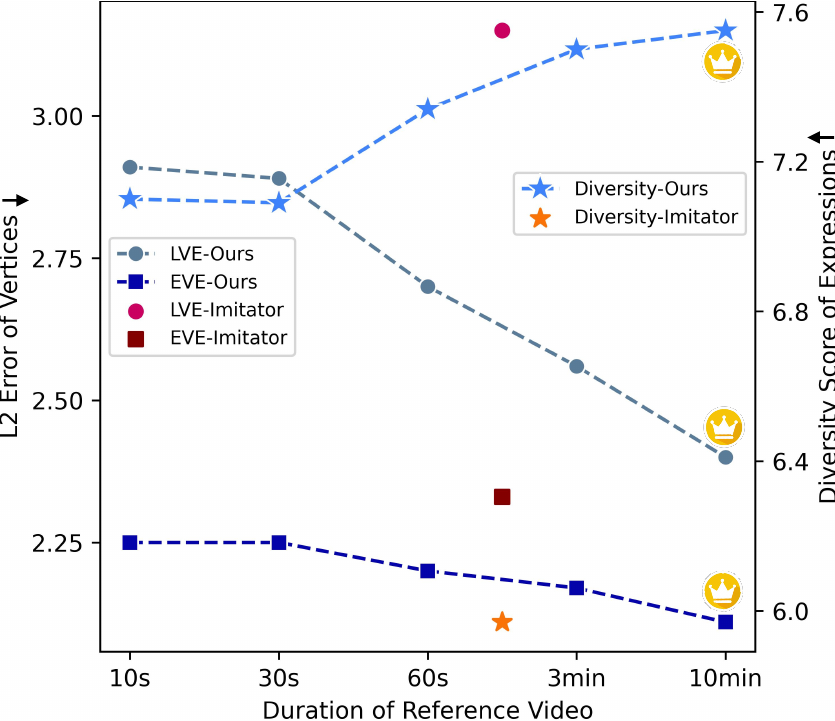}
    \caption{}
    \label{fig:pf_vs_data}
  \end{subfigure}
  \centering
  \begin{subfigure}{0.48\linewidth}
    \centering
    \includegraphics[scale=0.29]{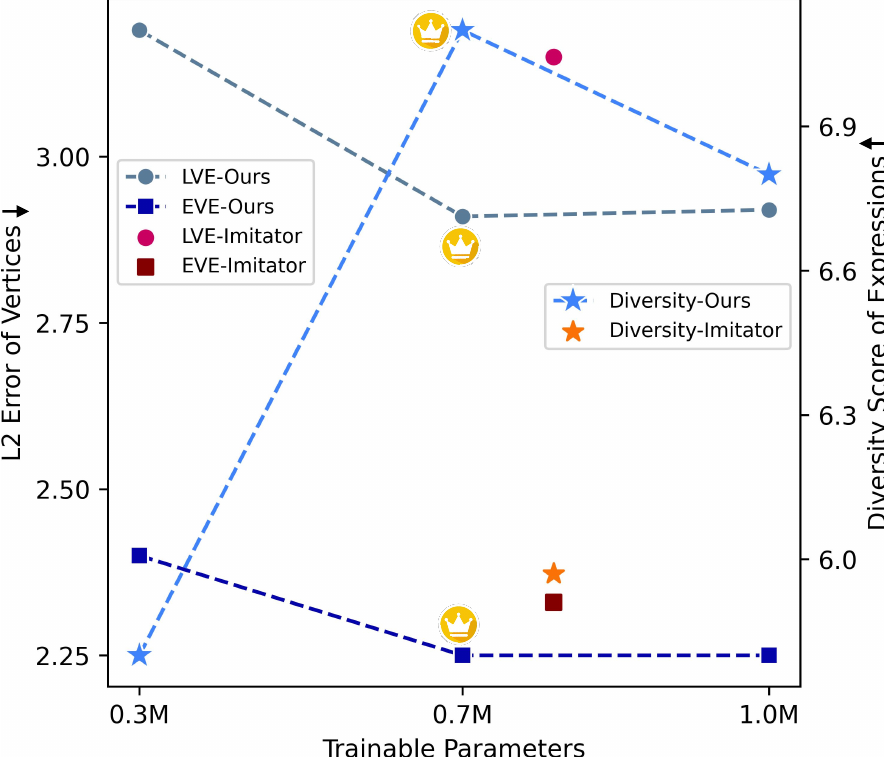}
    \caption{}
    \label{fig:pf_vs_param}
  \end{subfigure}
  \caption{Expression adapter analysis. (a) Performance vs data amount. (b) Performance vs adaptation parameter amount. {Lower L2 error of vertices and higher diversity score of expressions indicate better performance. The settings with the best performance are annotated with the yellow crowns.}}
  \label{fig:exp_ana}
\end{figure}
\begin{table}[t]    
    \centering
    \caption{Performance of MoLoRA and LoRA with different rank size settings. Values with the same \uline{single} or \uwave{wavy} underlines have approximately the same number of trainable parameters and can be compared.}
    \label{tab:ablation_molora}
    \begin{tabular}{lccccc}
        \toprule
        Setting   & Rank Size &\#Params & EVE$\downarrow$  & LVE$\downarrow$ &Div-{$\boldsymbol{\delta}$}$\uparrow$ \\ 
        \midrule
        \multicolumn{1}{l}{\multirow{3}{*}{MoLoRA}} & \textbf{\uwave{[4,8,16,32]}}  & {\textbf{\uwave{0.7M}}} & \textbf{\uwave{2.25}} & \textbf{\uwave{2.91}} &\textbf{\uwave{7.10}}      \\
        \multicolumn{1}{l}{}   & \uline{[4,32]} & {\uline{0.4M}} & \uline{2.32}  &\uline{3.07}  &\uline{6.52}      \\
        \multicolumn{1}{l}{} & [8,16] & {0.3M} & 2.36  & 3.10 & 6.21      \\ \hline
        \multicolumn{1}{l}{\multirow{3}{*}{LoRA}}      & 128 & {1.5M} & 2.29  & 3.00  & 6.34      \\
        \multicolumn{1}{l}{} & \uwave{64} & {\uwave{0.7M}}  & \uwave{2.38}  & \uwave{3.11}  & \uwave{5.80} \\
        \multicolumn{1}{l}{} & \uline{36} & {\uline{0.4M}}  & \uline{2.42}  & \uline{3.07}  & \uline{5.95}  \\
        \bottomrule
    \end{tabular}
\end{table}

\subsection{Approach Analysis}
Our approach involves multiple modules and training settings, thus we conduct extensive analysis to show the effectiveness of these configures.

\textbf{Ablation Study for Expression Adapter.} For expression adapter, we separately remove the following modules to verify their necessity. 1) As shown in Tab. \ref{tab:table2}, the removal of the style encoder leads to a drop of vertices errors and diversity metrics by a large margin. It suggests that the style encoder provides essential details which cannot be predicted from speech. 2) The absence of identity also causes a slight performance drop since the expression and identity parameters are not fully disentangled \cite{flame_2017,deca_2021}. The identity parameters also contain person-specific talking styles.
3) We fine-tune the full expression adapter instead of updating MoLoRA parameters. It can be observed that the performance significantly decreases since fine-tuning with few data damages the generalization ability of the pre-trained model.

\begin{figure}[t]
  \centering
  \begin{subfigure}{0.48\linewidth}
    \centering
    \includegraphics[scale=0.15]{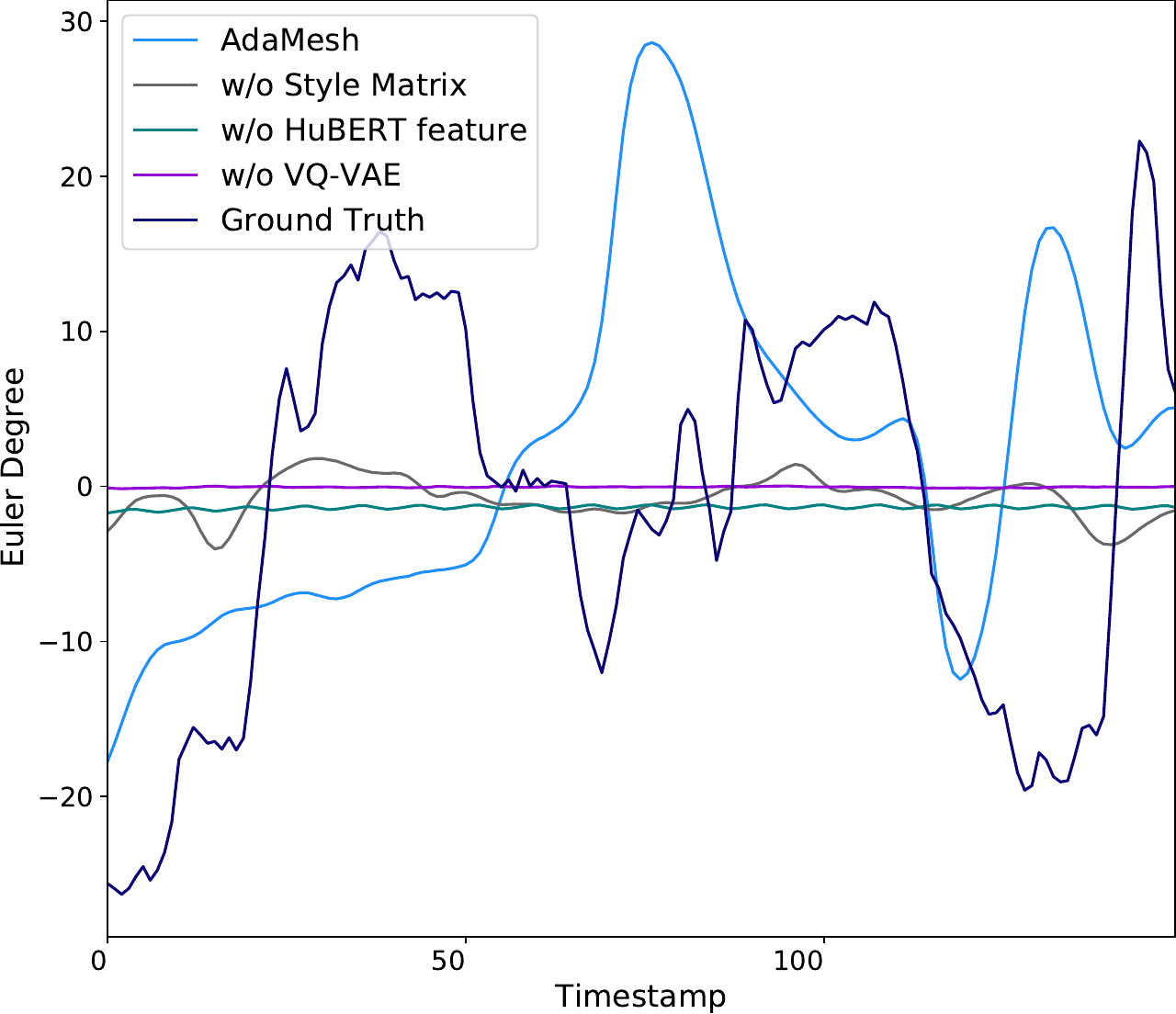}
    \caption{}
    \label{fig:eulers}
  \end{subfigure}
  \centering
  \begin{subfigure}{0.48\linewidth}
    \centering
    \includegraphics[scale=0.3]{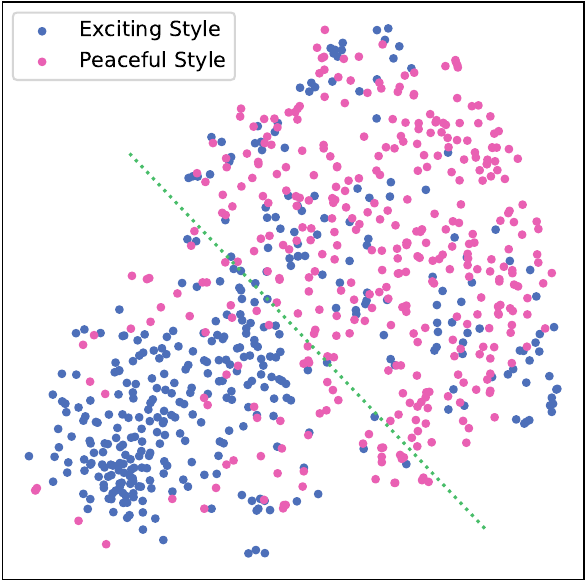}
    \caption{}
    \label{fig:tsne}
  \end{subfigure}
  \caption{Pose adapter analysis. (a) Yaw degree curves for head poses. (b) t-SNE visualization of
the semantic-aware pose style matrices.}
  \label{fig:pose_ana}
\end{figure}

\textbf{Ablation Study for Pose Adapter.} For pose adapter, we test three settings in Tab.\ref{tab: abla_pose}. 1) The retrieval of the pose style matrix is removed by randomly selecting a pose style embedding in the training data for inference. All metrics drop significantly due to the unawareness of the reference's personalized talking style. 2) w/o HuBERT feature, where only the mel spectrogram is utilized for the speech representation. The LSD and FSD scores show that HuBERT features contain semantic information, which is vital to the diverse and meaningful head pose generation. 3) After removing VQ-VAE, the model directly predicts the continuous poses. The diversity scores drop to near zero. It suggests that VQ-VAE provides realistic quantized codes which is easier to predict.
To further investigate the efficacy, we plot the yaw degree of head poses in Fig. \ref{fig:eulers}. The pose curve of AdaMesh has the largest variation, which is closer to the ground truth. It confirms the results in quantitative and qualitative experiments. We also utilize t-SNE \cite{tsne_2008}  to visualize the pose style matrix for the selected exciting and peaceful head poses. The visualization result shows that different styles are clustered into different groups. {This proves that the semantic-aware pose style matrix effectively aggregates and compresses different pose styles, thereby making the representation of pose styles not isolated and discrete, but continuous and meaningful. A similar example is that the inputs of BERT \cite{bert_2019} are one-hot discrete word embeddings, but the model learns meaningful representations with context-sensitive semantics.}

\begin{figure}[t]
    \centering
    \footnotesize
    \includegraphics[width=0.42\textwidth]{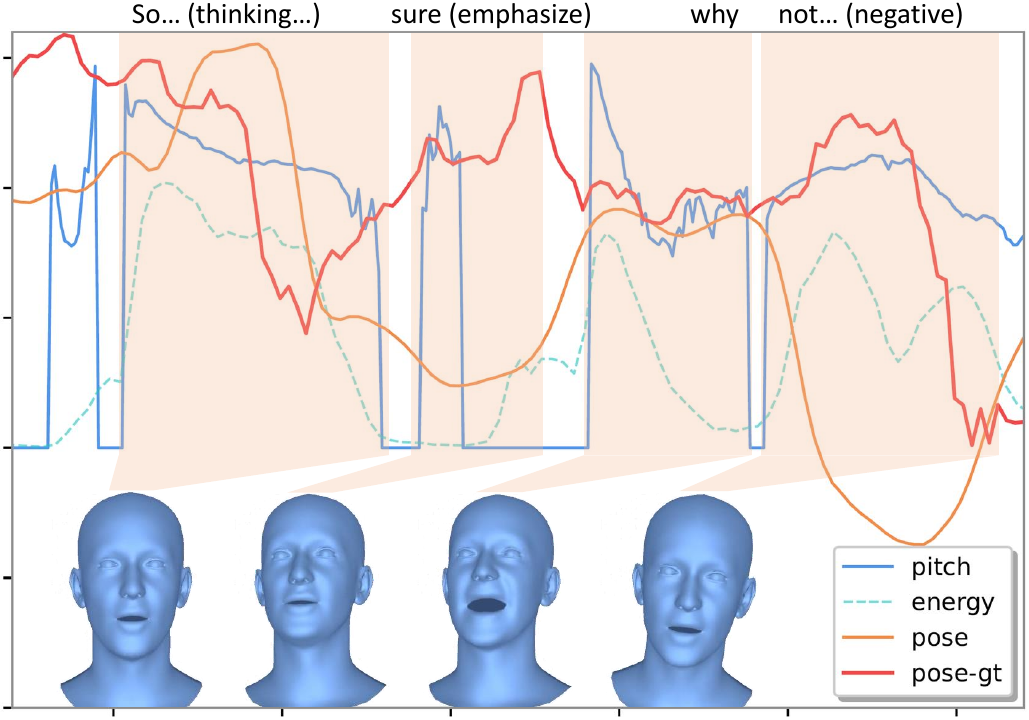}
    \caption{Correlation of speech features and head pose. Yaw degree is plotted as the head pose.}
    \label{fig: semantic_pose}
\end{figure}

\textbf{Amount of Adaptation Data.} We only analyze the impact of the adaptation data amounts on the expression adapter performance, since the pose adapter actually is a zero-shot model. As shown in Fig. \ref{fig:pf_vs_data},
points within the same color family represent the same method, while the same marker corresponds to the computation of the same metric. Thus the points with the same marker can be compared. The figure has two y-axises, and the legend next to each y-axis indicates what metric this y-axis represents. It can conclude that increasing the adaptation data amount can improve performance. AdaMesh achieves significantly better performance than Imitator with the same amount of adaptation data.
\begin{figure}[t]
    \centering
    \includegraphics[width=0.45\textwidth]{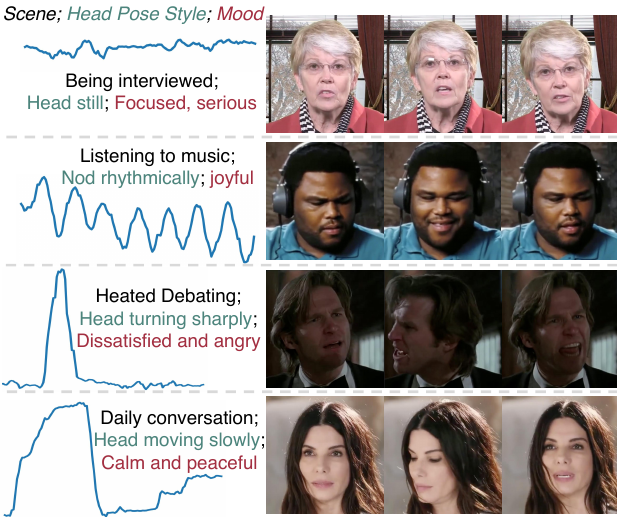}
    \caption{{Visualization of head pose styles for different speakers. Description, one of the head pose degrees and the corresponding video are demonstrated.}}
    \label{fig:headstyle_vis}
\end{figure}
\textbf{Amount of Adaption Parameters.} We estimate the impact of MoLoRA parameter amounts on the expression adapter performance. MoLoRA with different rank size combinations and the vanilla LoRA (the point in Fig. \ref{fig:pf_vs_param} with the maximum parameter amount) are compared. The point with the highest diversity score corresponds to the rank size combination of $[4, 8, 16, 32]$. This setting also achieves the lowest LVE and EVE scores. Furthermore, when optimizing roughly the same number of parameters, our approach outperforms Imitator in all metrics.

\textbf{Analysis on MoLoRA Rank Size.} Tab. \ref{tab:ablation_molora} shows the performance of the expression adapter with different rank sizes. It shows that increasing the rank size tends to yield better performance. The rank size combination of [4,8,16,32] for MoLoRA achieves the best performance. 
We compare with the vanilla LoRA \cite{lora_2022} of different rank sizes. 
{It is worth noting that with the same number of trainable parameters, MoLoRA achieves better results than the vanilla LoRA, even surpassing the LoRA with the rank size of 128. This indicates that the design of MoLoRA, which combines LoRAs of different rank sizes, enhance model performance more effectively than increasing the number of trainable parameters.} 

\textbf{Correlation between Speech and Head Poses.} Due to the subtle nature of head movements, it's challenging to observe their semantic correlations, especially when represented in a mesh format. Thus, we plot the speech features and the head pose curves in Fig. \ref{fig: semantic_pose}. The predicted and ground truth head poses are different in absolute value due to the one-to-many mapping between speech and pose. It is noticeable that significant changes in head poses correspond with variations in pitch and energy in speech (representing shifts in emphasis or emotion), demonstrating a positive correlation. Head poses distinctly change when the speech content involves emphasis or negation, indicating that the generated head poses contain semantic information, which is consistent with previous works \cite{jia2013headaf, audio2head2021, headtmm_2023}. The significant pose curve fluctuations on spoken words for the predicted head pose and the ground truth validates the reenactment of the reference style. Additionally, unstable head movements in the training dataset do not affect the generation of head poses, since the VQ-VAE, trained on a large dataset, effectively filtered out these unstable jitters. This can be confirmed by the user study.

{\textbf{Visualization of Specific Head Pose Styles.} As shown in Fig. \ref{fig:headstyle_vis}, we conduct a clustering analysis similar to Fig. \ref{fig:tsne} on the database for all pose style matrices and visualize some of the cluster centers and outliers. It demonstrates that different individuals have different head pose styles. By observing the corresponding ground-truth video and audio, we find that the head pose styles are influenced by multiple factors such as the scenes in which the speakers are situated, their personalities, and moods. This verifies the rationale for the targeted modeling of head pose styles.}
\begin{figure}[t]
    \centering
    \includegraphics[width=0.46\textwidth]{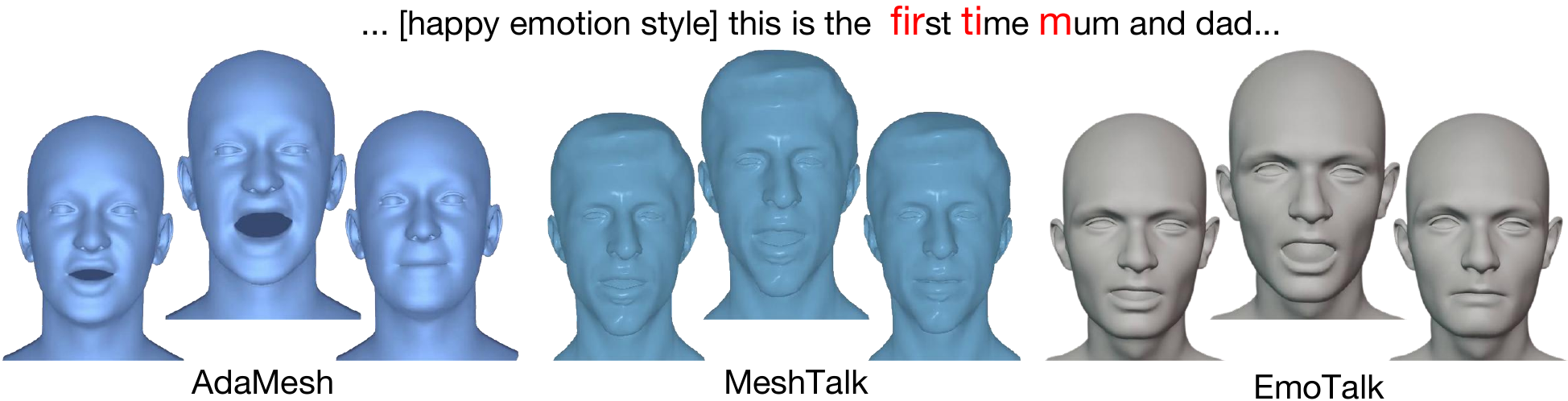}
    \caption{Visual comparison with methods of different topologies. Emotion can be observed on AdaMesh results.}
    \label{fig: topology}
\end{figure}
\begin{figure}[t]
    \centering
    \footnotesize
    \includegraphics[width=0.47\textwidth]{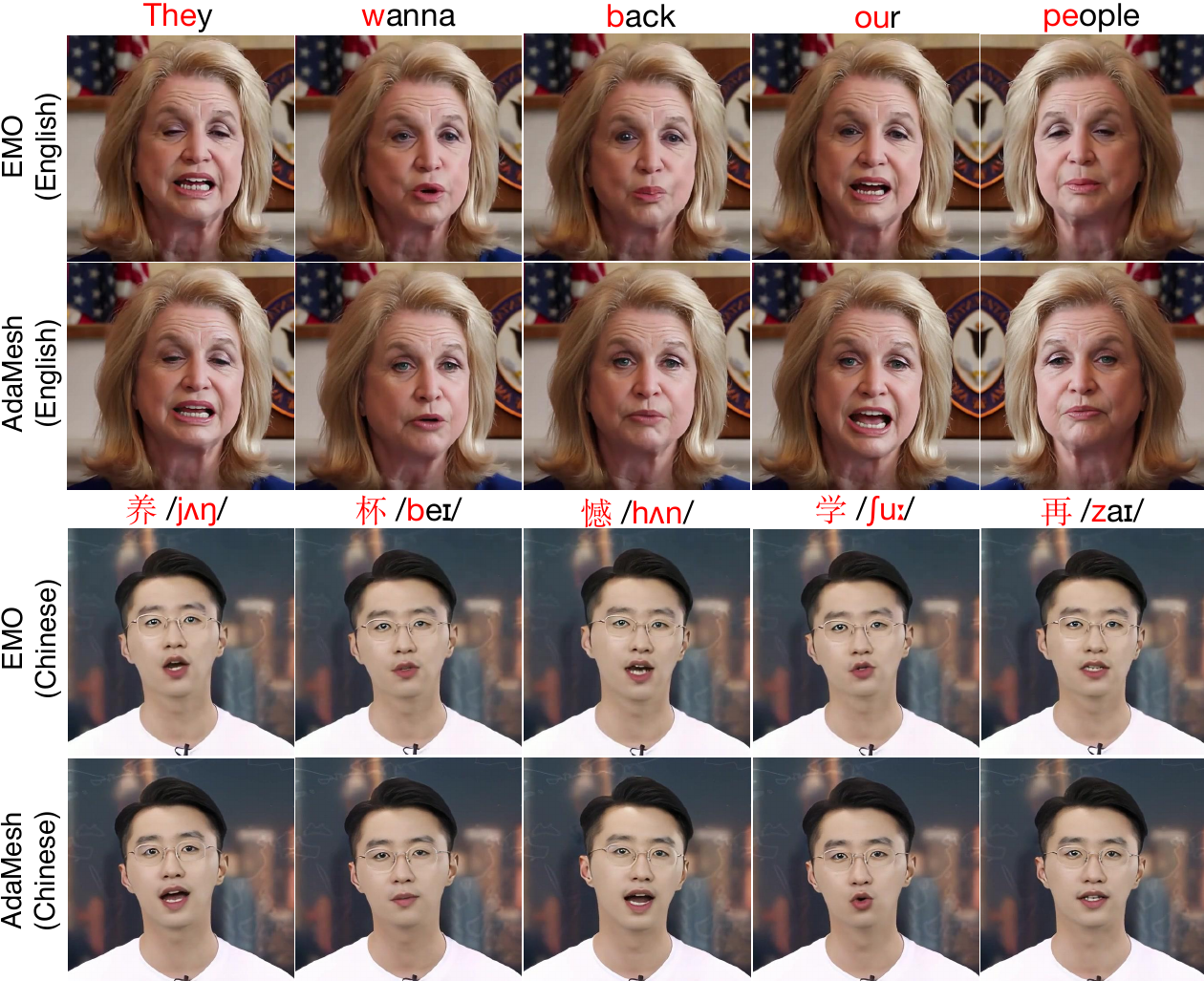}
    \caption{Qualitative comparison with EMO \cite{emo_2024} on the portrait-realistic talking face generation task. English and Chinese results are given separately. We annotate the closest equivalent English phonetic symbols next to the Chinese pronunciation for better observation.}
    \label{fig: render}
\end{figure}

\subsection{More Demonstration}
\textbf{Comparison with Other Topologies.} In Fig. \ref{fig: topology}, AdaMesh is compared with methods of different topologies \cite{meshtalk_2021, emotalk} that cannot be easily converted to FLAME format \cite{flame_2017}. AdaMesh still achieves better lip sync and richer facial expressions.

\textbf{Portrait-Realistic Talking Face Generation.} In Fig.\ref{fig: render}, we further show that our model can generate coherent photo-realistic video by employing a simple conditional GAN-based person-specific image renderer \cite{nvp_2020,vast_2023}. Expressions and head poses are predicted by the proposed AdaMesh. Then the image renderer takes the rasterized 2D mesh sequence and the identity image sequence as inputs and synthesizes a portrait video, which matches the expression details of the mesh sequence and the appearance of the identity sequence.
We provide the results of a state-of-the-art talking face method EMO \cite{emo_2024} as a comparison to AdaMesh. EMO is an end-to-end method that directly predicts image variation from speech without any intermediate face representation, thus achieving highly synchronized lip movements. AdaMesh gains comparable lip synchronization performance with EMO. This suggests that our
approach can predict accurate-enough expressions compared with the end-to-end method and can also serve as a conduit for photo-realistic avatars. 

{\textbf{Generation Efficiency.} When inference on Nvidia A10 GPU, it takes 10ms from inputting speech, predicting facial parameters, to generating meshes. Utilizing the renderer to synthesize portrait-realistic images will cost 9ms. Our method can be employed in the realistic human live-streaming scenario with a rapid inference speed at 48fps.}
\section{Conclusion}
This paper proposes AdaMesh for speech-driven 3D facial animation, which aims to capture the personalized talking style from the given reference video, to generate rich facial expressions and diverse head poses. 
Our central insight is to separately devise suitable adaptation strategies for facial expressions and head poses. The MoLoRA and retrieval adaptation strategies are in line with the characteristics of these two types of data.
The proposed approach successfully solves the catastrophic forgetting and overfit problems for facial expression modeling and the averaged generation problem for head pose modeling under the circumstance of few data adaptation. Extensive experimental results and approach analysis prove that AdaMesh archives high-quality and efficient style adaptation, and outperforms other state-of-the-art methods.

\normalem
\bibliography{main}

\begin{thebibliography}{10}
\providecommand{\url}[1]{#1}
\csname url@samestyle\endcsname
\providecommand{\newblock}{\relax}
\providecommand{\bibinfo}[2]{#2}
\providecommand{\BIBentrySTDinterwordspacing}{\spaceskip=0pt\relax}
\providecommand{\BIBentryALTinterwordstretchfactor}{4}
\providecommand{\BIBentryALTinterwordspacing}{\spaceskip=\fontdimen2\font plus
\BIBentryALTinterwordstretchfactor\fontdimen3\font minus \fontdimen4\font\relax}
\providecommand{\BIBforeignlanguage}[2]{{%
\expandafter\ifx\csname l@#1\endcsname\relax
\typeout{** WARNING: IEEEtran.bst: No hyphenation pattern has been}%
\typeout{** loaded for the language `#1'. Using the pattern for}%
\typeout{** the default language instead.}%
\else
\language=\csname l@#1\endcsname
\fi
#2}}
\providecommand{\BIBdecl}{\relax}
\BIBdecl

\bibitem{voca_2019}
D.~Cudeiro, T.~Bolkart, C.~Laidlaw, A.~Ranjan, and M.~J. Black, ``Capture, learning, and synthesis of 3d speaking styles,'' in \emph{2019 IEEE/CVF Conference on Computer Vision and Pattern Recognition (CVPR)}, 2019, pp. 10\,093--10\,103.

\bibitem{meshtalk_2021}
A.~Richard, M.~Zollhofer, Y.~Wen, F.~de~la Torre, and Y.~Sheikh, ``Meshtalk: 3d face animation from speech using cross-modality disentanglement,'' in \emph{2021 IEEE/CVF International Conference on Computer Vision (ICCV)}, 2021, pp. 1153--1162.

\bibitem{faceformer_2022}
Y.~Fan, Z.~Lin, J.~Saito, W.~Wang, and T.~Komura, ``Faceformer: Speech-driven 3d facial animation with transformers,'' in \emph{CVPR}, 2022, pp. 18\,749--18\,758.

\bibitem{codetalker_2023}
J.~Xing, M.~Xia, Y.~Zhang, X.~Cun, J.~Wang, and T.~Wong, ``Codetalker: Speech-driven 3d facial animation with discrete motion prior,'' in \emph{CVPR}, 2023, pp. 12\,780--12\,790.

\bibitem{sdnerf_2024}
S.~Shen, W.~Li, X.~Huang, Z.~Zhu, J.~Zhou, and J.~Lu, ``Sd-nerf: Towards lifelike talking head animation via spatially-adaptive dual-driven nerfs,'' \emph{IEEE Transactions on Multimedia}, vol.~26, pp. 3221--3234, 2024.

\bibitem{mimic_2024}
H.~Fu, Z.~Wang, K.~Gong, K.~Wang, T.~Chen, H.~Li, H.~Zeng, and W.~Kang, ``Mimic: Speaking style disentanglement for speech-driven 3d facial animation,'' \emph{Proceedings of the AAAI Conference on Artificial Intelligence}, vol.~38, no.~2, pp. 1770--1777, Mar. 2024.

\bibitem{diffposetalk_2024}
\BIBentryALTinterwordspacing
Z.~Sun, T.~Lv, S.~Ye, M.~Lin, J.~Sheng, Y.-H. Wen, M.~Yu, and Y.-J. Liu, ``Diffposetalk: Speech-driven stylistic 3d facial animation and head pose generation via diffusion models,'' \emph{ACM Trans. Graph.}, vol.~43, no.~4, jul 2024. [Online]. Available: \url{https://doi.org/10.1145/3658221}
\BIBentrySTDinterwordspacing

\bibitem{geneface_2023}
Z.~Ye, Z.~Jiang, Y.~Ren, J.~Liu, J.~He, and Z.~Zhao, ``Geneface: Generalized and high-fidelity audio-driven 3d talking face synthesis,'' in \emph{Proceedings of the International Conference on Learning Representations (ICLR)}, 2023.

\bibitem{imitator_2023}
B.~Thambiraja, I.~Habibie, S.~Aliakbarian, D.~Cosker, C.~Theobalt, and J.~Thies, ``Imitator: Personalized speech-driven 3d facial animation,'' in \emph{ICCV}, 2023.

\bibitem{laughtalk_2024}
K.~Sung-Bin, L.~Hyun, D.~h. Hong, S.~Nam, J.~Ju, and T.-H. Oh, ``Laughtalk: Expressive 3d talking head generation with laughter,'' in \emph{IEEE/CVF Winter Conference on Applications of Computer Vision (WACV)}, 2024.

\bibitem{catastrophy_2023}
Y.~Zhai, S.~Tong, X.~Li, M.~Cai, Q.~Qu, Y.~J. Lee, and Y.~Ma, ``Investigating the catastrophic forgetting in multimodal large language models,'' 2023.

\bibitem{fewshot_zhang_2022}
W.~Zhang, L.~Shen, W.~Zhang, and C.-S. Foo, ``Few-shot adaptation of pre-trained networks for domain shift,'' in \emph{Proceedings of International Joint Conference on Artificial Intelligence (IJCAI)}, 2022, pp. 1665--1671.

\bibitem{moglow_2020}
G.~E. Henter, S.~Alexanderson, and J.~Beskow, ``Mo{G}low: {P}robabilistic and controllable motion synthesis using normalising flows,'' \emph{ACM Transactions on Graphics}, vol.~39, no.~4, pp. 236:1--236:14, 2020.

\bibitem{facial_2021}
C.~Zhang, Y.~Zhao, Y.~Huang, M.~Zeng, S.~Ni, M.~Budagavi, and X.~Guo, ``Facial: Synthesizing dynamic talking face with implicit attribute learning,'' in \emph{ICCV}, 2021, pp. 3867--3876.

\bibitem{lora_2022}
E.~J. Hu, Y.~Shen, P.~Wallis, Z.~Allen-Zhu, Y.~Li, S.~Wang, L.~Wang, and W.~Chen, ``Lo{RA}: Low-rank adaptation of large language models,'' in \emph{ICLR}, 2022.

\bibitem{vqvae_2018}
A.~van~den Oord, O.~Vinyals, and K.~Kavukcuoglu, ``Neural discrete representation learning,'' 2018.

\bibitem{transformer_2017}
A.~Vaswani, N.~Shazeer, N.~Parmar, J.~Uszkoreit, L.~Jones, A.~N. Gomez, L.~u. Kaiser, and I.~Polosukhin, ``Attention is all you need,'' in \emph{Advances in Neural Information Processing Systems}, vol.~30, 2017.

\bibitem{biwi_2010}
G.~Fanelli, J.~Gall, H.~Romsdorfer, T.~Weise, and L.~Van~Gool, ``A 3-d audio-visual corpus of affective communication,'' \emph{IEEE Transactions on Multimedia}, vol.~12, no.~6, pp. 591--598, 2010.

\bibitem{flame_2017}
T.~Li, T.~Bolkart, M.~J. Black, H.~Li, and J.~Romero, ``Learning a model of facial shape and expression from 4d scans,'' \emph{ACM Trans. Graph.}, vol.~36, no.~6, 2017.

\bibitem{emote}
R.~Daněček, K.~Chhatre, S.~Tripathi, Y.~Wen, M.~J. Black, and T.~Bolkart, ``Emotional speech-driven animation with content-emotion disentanglement,'' 2023.

\bibitem{transformers2a}
L.~Chen, Z.~Wu, J.~Ling, R.~Li, X.~Tan, and S.~Zhao, ``Transformer-s2a: Robust and efficient speech-to-animation,'' in \emph{IEEE International Conference on Acoustics, Speech and Signal Processing (ICASSP)}, 2022, pp. 7247--7251.

\bibitem{emotalk}
Z.~Peng, H.~Wu, Z.~Song, H.~Xu, X.~Zhu, J.~He, H.~Liu, and Z.~Fan, ``Emotalk: Speech-driven emotional disentanglement for 3d face animation,'' in \emph{IEEE/CVF International Conference on Computer Vision (ICCV)}, 2023, pp. 20\,630--20\,640.

\bibitem{nvp_2020}
J.~Thies, M.~Elgharib, A.~Tewari, C.~Theobalt, and M.~Nie{\ss}ner, ``Neural voice puppetry: Audio-driven facial reenactment,'' \emph{ECCV 2020}, 2020.

\bibitem{vast_2023}
L.~Chen, Z.~Wu, R.~Li, W.~Bao, J.~Ling, X.~Tan, and S.~Zhao, ``Vast: Vivify your talking avatar via zero-shot expressive facial style transfer,'' in \emph{Proceedings of the IEEE/CVF International Conference on Computer Vision Workshop (ICCV-W)}, 2023.

\bibitem{taming_2023}
L.~Zhu, X.~Liu, X.~Liu, R.~Qian, Z.~Liu, and L.~Yu, ``Taming diffusion models for audio-driven co-speech gesture generation,'' in \emph{Proceedings of the IEEE/CVF Conference on Computer Vision and Pattern Recognition}, 2023, pp. 10\,544--10\,553.

\bibitem{zhanghead2007}
S.~Zhang, Z.~Wu, H.~M. Meng, and L.~Cai, ``Head movement synthesis based on semantic and prosodic features for a chinese expressive avatar,'' in \emph{IEEE International Conference on Acoustics, Speech and Signal Processing}, vol.~4, 2007, pp. IV--837--IV--840.

\bibitem{jia2013headaf}
J.~Jia, Z.~Wu, S.~Zhang, H.~M. Meng, and L.~Cai, ``Head and facial gestures synthesis using pad model for an expressive talking avatar,'' \emph{Multimedia Tools and Applications}, vol.~73, pp. 439 -- 461, 2013.

\bibitem{audio2head2021}
S.~Wang, L.~Li, Y.~Ding, C.~Fan, and X.~Yu, ``Audio2head: Audio-driven one-shot talking-head generation with natural head motion,'' in \emph{Proceedings of the Thirtieth International Joint Conference on Artificial Intelligence}, 2021, pp. 1098--1105.

\bibitem{headtmm_2023}
R.~Yi, Z.~Ye, Z.~Sun, J.~Zhang, G.~Zhang, P.~Wan, H.~Bao, and Y.-J. Liu, ``Predicting personalized head movement from short video and speech signal,'' \emph{IEEE Transactions on Multimedia}, vol.~25, pp. 6315--6328, 2023.

\bibitem{sadtalker_2023}
W.~Zhang, X.~Cun, X.~Wang, Y.~Zhang, X.~Shen, Y.~Guo, Y.~Shan, and F.~Wang, ``Sadtalker: Learning realistic 3d motion coefficients for stylized audio-driven single image talking face animation,'' in \emph{2023 IEEE/CVF Conference on Computer Vision and Pattern Recognition (CVPR)}, 2023, pp. 8652--8661.

\bibitem{luhead2024}
J.~Lu and H.~Shimodaira, ``Speech-driven head motion generation from waveforms,'' \emph{Speech Communication}, vol. 159, p. 103056, 2024.

\bibitem{adapter_2017}
S.-A. Rebuffi, H.~Bilen, and A.~Vedaldi, ``Learning multiple visual domains with residual adapters,'' in \emph{Advances in Neural Information Processing Systems}, 2017.

\bibitem{prefix_2021}
X.~L. Li and P.~Liang, ``Prefix-tuning: Optimizing continuous prompts for generation,'' in \emph{Proceedings of the 59th Annual Meeting of the Association for Computational Linguistics}, 2021, pp. 4582--4597.

\bibitem{sd_2022}
R.~Rombach, A.~Blattmann, D.~Lorenz, P.~Esser, and B.~Ommer, ``High-resolution image synthesis with latent diffusion models,'' in \emph{CVPR}, 2022.

\bibitem{conformer_2021}
S.~Li, M.~Xu, and X.-L. Zhang, ``Conformer-based end-to-end speech recognition with rotary position embedding,'' in \emph{2021 Asia-Pacific Signal and Information Processing Association Annual Summit and Conference (APSIPA ASC)}, 2021, pp. 443--447.

\bibitem{spectre_2023}
P.~P. Filntisis, G.~Retsinas, F.~Paraperas-Papantoniou, A.~Katsamanis, A.~Roussos, and P.~Maragos, ``Spectre: Visual speech-informed perceptual 3d facial expression reconstruction from videos,'' in \emph{2023 IEEE/CVF Conference on Computer Vision and Pattern Recognition Workshops (CVPRW)}, 2023, pp. 5745--5755.

\bibitem{hubert_2021}
W.-N. Hsu, B.~Bolte, Y.-H.~H. Tsai, K.~Lakhotia, R.~Salakhutdinov, and A.~Mohamed, ``Hubert: Self-supervised speech representation learning by masked prediction of hidden units,'' 2021.

\bibitem{deca_2021}
Y.~Feng, H.~Feng, M.~J. Black, and T.~Bolkart, ``Learning an animatable detailed 3d face model from in-the-wild images,'' \emph{ACM Trans. Graph.}, vol.~40, no.~4, 2021.

\bibitem{adaspeech_2021}
M.~Chen, X.~Tan, B.~Li, Y.~Liu, T.~Qin, S.~Zhao, and T.-Y. Liu, ``Adaspeech: Adaptive text to speech for custom voice,'' in \emph{Proceedings of the International Conference on Learning Representations (ICLR)}, 2021.

\bibitem{soundstream_2021}
N.~Zeghidour, A.~Luebs, A.~Omran, J.~Skoglund, and M.~Tagliasacchi, ``Soundstream: An end-to-end neural audio codec,'' 2021.

\bibitem{autoencoder_2022}
\BIBentryALTinterwordspacing
D.~P. Kingma and M.~Welling, ``Auto-encoding variational bayes,'' 2022. [Online]. Available: \url{https://arxiv.org/abs/1312.6114}
\BIBentrySTDinterwordspacing

\bibitem{talklora_2024}
\BIBentryALTinterwordspacing
J.~Saunders and V.~Namboodiri, ``Talklora: Low-rank adaptation for speech-driven animation,'' 2024. [Online]. Available: \url{https://arxiv.org/abs/2408.13714}
\BIBentrySTDinterwordspacing

\bibitem{metaloraanimation_2024}
\BIBentryALTinterwordspacing
X.~Zhou, F.~Li, Z.~Peng, K.~Wu, J.~He, B.~Qin, Z.~Fan, and H.~Liu, ``Meta-learning empowered meta-face: Personalized speaking style adaptation for audio-driven 3d talking face animation,'' 2024. [Online]. Available: \url{https://arxiv.org/abs/2408.09357}
\BIBentrySTDinterwordspacing

\bibitem{relora_2023}
V.~Lialin, N.~Shivagunde, S.~Muckatira, and A.~Rumshisky, ``Stack more layers differently: High-rank training through low-rank updates,'' 2023.

\bibitem{bailando_2022}
L.~Siyao, W.~Yu, T.~Gu, C.~Lin, Q.~Wang, C.~Qian, C.~C. Loy, and Z.~Liu, ``Bailando: 3d dance generation via actor-critic gpt with choreographic memory,'' in \emph{CVPR}, 2022.

\bibitem{learn2listen_2022}
E.~Ng, H.~Joo, L.~Hu, H.~Li, T.~Darrell, A.~Kanazawa, and S.~Ginosar, ``Learning to listen: Modeling non-deterministic dyadic facial motion,'' in \emph{CVPR}, 2022, pp. 20\,395--20\,405.

\bibitem{synobama_2017}
S.~Suwajanakorn, S.~M. Seitz, and I.~Kemelmacher-Shlizerman, ``Synthesizing obama: Learning lip sync from audio,'' \emph{ACM Trans. Graph.}, vol.~36, no.~4, 2017.

\bibitem{vox2_2018}
J.~S. Chung, A.~Nagrani, and A.~Zisserman, ``{VoxCeleb2: Deep Speaker Recognition},'' in \emph{Proc. Interspeech 2018}, 2018, pp. 1086--1090.

\bibitem{mead_2020}
K.~Wang, Q.~Wu, L.~Song, Z.~Yang, W.~Wu, C.~Qian, R.~He, Y.~Qiao, and C.~C. Loy, ``Mead: A large-scale audio-visual dataset for emotional talking-face generation,'' in \emph{ECCV}, 2020.

\bibitem{deep3d_2019}
Y.~Deng, J.~Yang, S.~Xu, D.~Chen, Y.~Jia, and X.~Tong, ``Accurate 3d face reconstruction with weakly-supervised learning: From single image to image set,'' in \emph{IEEE Computer Vision and Pattern Recognition Workshops}, 2019.

\bibitem{tsne_2008}
L.~Van~der Maaten and G.~Hinton, ``Visualizing data using {t-SNE}.'' \emph{Journal of machine learning research (JMLR)}, vol.~9, no.~11, pp. 2579--2605, 2008.

\bibitem{bert_2019}
J.~Devlin, M.-W. Chang, K.~Lee, and K.~Toutanova, ``{BERT}: Pre-training of deep bidirectional transformers for language understanding,'' in \emph{Proceedings of the 2019 Conference of the North {A}merican Chapter of the Association for Computational Linguistics: Human Language Technologies, Volume 1 (Long and Short Papers)}, 2019, pp. 4171--4186.

\bibitem{emo_2024}
L.~Tian, Q.~Wang, B.~Zhang, and L.~Bo, ``Emo: Emote portrait alive -- generating expressive portrait videos with audio2video diffusion model under weak conditions,'' 2024.

\end{thebibliography}
\bibliographystyle{IEEEtran}


 




\vfill

\end{document}